\documentclass[a4paper,conference]{IEEEtran}
\ifCLASSINFOpdf
   \usepackage[pdftex]{graphicx}
\else
   \usepackage[dvips]{graphicx}
\fi
\usepackage{array}
\usepackage{xcolor}

 \usepackage[caption=false,font=normalsize,labelfont=sf,textfont=sf]{subfig}

\usepackage{algorithm}
\usepackage[noend]{algpseudocode}

\makeatletter
\def\BState{\State\hskip-\ALG@thistlm}
\makeatother
\usepackage{xcolor}
\usepackage{colortbl}
\usepackage{mathtools}
\DeclarePairedDelimiter\ceil{\lceil}{\rceil}

\begin{document}

\title{Recognizing American Sign Language Nonmanual Signal Grammar Errors in Continuous Videos}

\author{
\IEEEauthorblockN{Elahe Vahdani, Longlong Jing, and Yingli Tian*}
\IEEEauthorblockA{The Graduate Center and The City College,\\ 
The City University of New York,
New York, NY, USA. \\
\{evahdani,~ljing\}@gradcenter.cuny.edu\\
ytian@ccny.cuny.edu (*Corresponding author)\\
}
\and
\IEEEauthorblockN{Matt Huenerfauth}
\IEEEauthorblockA{Golisano College of Computing and Information Sciences\\
The Rochester Institute of Technology\\
Rochester, NY, USA.\\ 
matt.huenerfauth@rit.edu}\\
}

\maketitle

\begin{abstract}
As part of the development of an educational tool that can help students achieve fluency in American Sign Language (ASL) through independent and interactive practice with immediate feedback, this paper introduces a near real-time system to recognize grammatical errors in continuous signing videos without necessarily identifying the entire sequence of signs. Our system automatically recognizes if a performance of ASL sentences contains grammatical errors made by ASL students. We first recognize the ASL grammatical elements including both manual gestures and nonmanual signals independently from multiple modalities (i.e. hand gestures, facial expressions, and head movements) by 3D-ResNet networks. Then the temporal boundaries of grammatical elements from different modalities are examined to detect ASL grammatical mistakes by using a sliding window-based approach. We have collected a dataset of continuous sign language, \textbf{ASL-HW-RGBD}, covering different aspects of ASL grammars for training and testing. Our system is able to recognize grammatical elements on ASL-HW-RGBD from manual gestures, facial expressions, and head movements and successfully detect 8 ASL grammatical mistakes.

\end{abstract}


\IEEEpeerreviewmaketitle

\section{INTRODUCTION} \label{section:intro}

\subsection{Motivation and Challenges}
American Sign Language (ASL) has become the 4th most studied language at U.S. colleges \cite{Furman10}. To assist ASL students practicing their signing skills, we design a near real-time deep learning-based system to process continuous signing videos and evaluate students' performance. Our system is able to recognize a set of ASL grammatical mistakes by processing multiple modalities including hand gestures, facial expressions, and head movements from continuous videos.

Compared to isolated sign language recognition \cite{jing2019recognizing}, continuous sign language recognition (CSLR) is more complicated because no temporal boundaries of signs are provided. Moreover, the transition movements between two consecutive signs are subtle and diverse and therefore hard to detect. Our goal is  to recognize grammatically important components and detect  possible mistakes without fully recognizing each individual word in continuous ASL videos.

ASL is a natural language with a distinct grammar structure from English, as illustrated by how questions may be formed in each language:  In English, \textit{``WH-Questions''} use words like \textit{Who, What, Where, When, Why}, typically at the beginning of a sentence, e.g. ``What did she buy yesterday?''  In ASL, WH-word may occur in other positions in the sentence, including at the end, e.g. ``SHE BUY YESTERDAY WHAT.'' The question is primarily indicated by a ``nonmanual signal'' during a WH-word or a longer span of the sentence.  Nonmanual signals may comprise several body movements such as head, shoulders, torso, eyebrows, eyeballs (gaze), eyelids, nose, mouth, tongue, cheeks, and chin \cite{wilbur2009effects}.  The nonmanual signal for WH-Questions consists of a furrowing of brows and a tilt of head forward.   In English, to change a declarative sentence to a \textit{``Yes-No-Question''}, one can change the order of words from ``She was there'' to ``Was she there?''.  In ASL, to express a \textit{``Yes-No-Question''}, rather than changing the word order, the question is indicated by a nonmanual signal (eyebrows raised and head tilted forward), especially at the end of the sentence \cite{perlmutter2011sign}.

\begin{figure}[!t]
  \includegraphics[width=\linewidth]{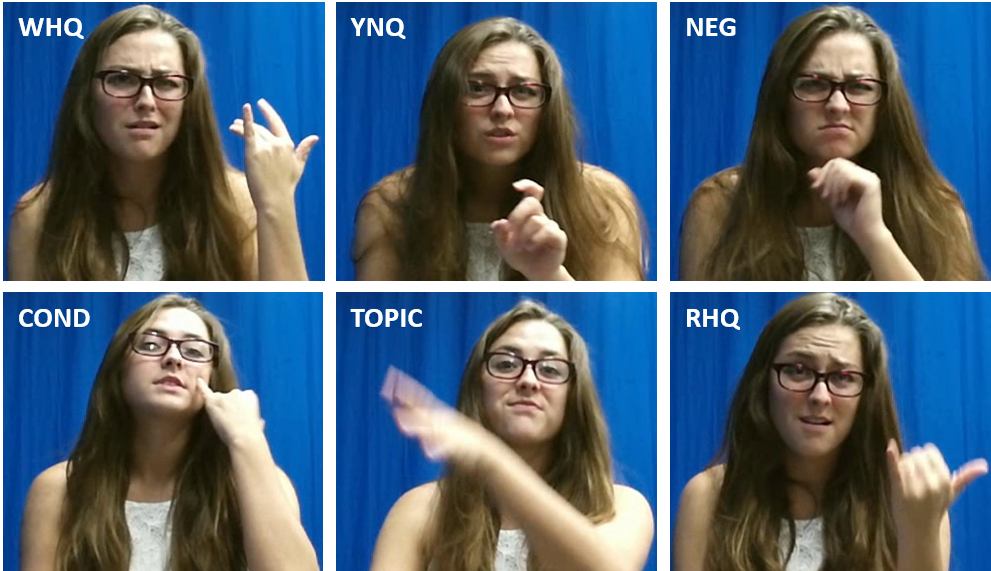}
  \caption{Examples of ASL grammatical elements from multimodalities including facial expressions, head movements, and hand gestures: WH-Question (WHQ), Yes-No-Question (YNQ), Negative (NEG), Conditional (COND), Topic, and Rhetorical Question (RHQ). }
 \vspace{-0.1in}
\label{fig:FE}
\end{figure}

Nonmanual signals are used to convey a variety of grammatical information in ASL, e.g. Topicalization, \textit{``Yes-No-Question''}, \textit{``WH-Question''}, etc.  Because they consist of movements of multiple body parts, our recognition framework is built upon multiple modalities to detect these ASL grammatical elements. Note that different nonmanual signals may share behavioral properties, increasing the recognition difficulty. For example, during \textit{``WH-Question''} and \textit{``Yes-No-Question''}, head tilts forward for both while eyebrows are furrowed during the former and raised during the latter, see Fig. \ref{fig:FE}. To address this challenge, we extract fine-grained features from different modalities. 

\subsection{ASL Linguistic Elements Recognized by Our System }
A grammatically correct ASL sentence requires  signer to synchronize manual signs (primarily consisting of gestures of the hands, see Table \ref{table:manual_gesture_list}) and nonmanual signals (primarily consisting of facial expressions and head movements, see Table \ref{table:FE_list}).  For instance, if someone performs a WH-question sentence in ASL without performing the WH-Question nonmanual signal during the WH-word, this would be ungrammatical.   
As another example, if someone performs a ``Negative'' word, e.g. NOT, NONE, etc., without also performing a ``Negative'' nonmanual signal at the same time, this would also be ungrammatical.  Thus, our system leverages these types of relationships between key classes of ASL manual signs and required ASL nonmanual signals to detect when the person in a video may be making an error. Since nonmanual signals often span multiple manual signs, and since movements of the head may require some onset/offset transitional time, many possible temporal alignments between the nonmanual and manual channels are possible, including various forms of overlap.  Thus, the error-detection rules in our system use a temporal threshold to determine whether a grammatically-required nonmanual signal has occurred at the correct time, relative to the timing of the manual signs. 

Table \ref{table:describ-errors} lists the ASL grammatical errors recognized in our system. We categorize them into two groups: 1) Lexical errors happen with absence of proper facial expressions and head movements when certain classes of manual signs are performed, 2) Timing errors indicate that a grammatically important nonmanual signal happened but was too far from the clause boundary.  This second group captures how signers performing ASL must produce certain nonmanual signals in temporal alignment with the beginning or end of sentences or clauses.  In our system, students may submit a video of longer passages of ASL consisting of multiple sentences; therefore, a subset of the inter-sign boundaries in the video will also be ``clause boundaries'' (when a word is the first or last in a clause).  Our system must therefore also detect clause boundaries throughout continuous videos. 

\begin{table}[!t]
\centering
\caption{Classes of Manual Signs Detected by Our System. }
\resizebox{0.5\textwidth}{!}{
\begin{tabular}{l l }
  \hline
  Gesture Type & Description and Instances \\
  \hline
  \hline
\cellcolor{gray!20} Conditional	&  \cellcolor{gray!20} IF-SUPPOSE \\
\cellcolor{gray!5} Negative	& \cellcolor{gray!5} NEVER, NONE, NOT, DON'T MIND \\
\cellcolor{gray!20} YNQ	& \cellcolor{gray!20} QUESTION MARK WIGGLE, QUESTION  \\
\cellcolor{gray!5} WHQ	& \cellcolor{gray!5} WHO, WHAT, WHERE, WHY, WHEN \\
\cellcolor{gray!20} Time & \cellcolor{gray!20} NOW, TODAY, ALWAYS, LAST-WEEK \\
\cellcolor{gray!5} Pointing	&  \cellcolor{gray!5} IX-HE-SHE-IT, I-ME, YOU \\
\cellcolor{gray!5} Fingerspelling	& \cellcolor{gray!5} Spelling out the letters by handshapes.\\
\cellcolor{gray!20} Clause Boundary	& \cellcolor{gray!20} Gap between two consecutive clauses.\\
\hline
\end{tabular}} 
\label{table:manual_gesture_list} 
\end{table}

\begin{table}[!t]
\centering
\caption{Classes of Nonmanual Signals Our System Detects.}
\resizebox{0.5\textwidth}{!}{
\begin{tabular}{l  l   l }
  \hline
  Face Class & Head Pose (Head Class) & Eyebrows, Eyes, Nose \\
  \hline
  \hline
\cellcolor{gray!5}Negative & \cellcolor{gray!5}Shaking side-to-side (1)  & \cellcolor{gray!5}Furrowed, Scrunched nose \\
\cellcolor{gray!20}YNQ	&  \cellcolor{gray!20}Tilted forward (2) &\cellcolor{gray!20}Raised \\
\cellcolor{gray!5}WHQ	&  \cellcolor{gray!5}Tilted forward (2) & \cellcolor{gray!5}Furrowed  \\
\cellcolor{gray!20}RHQ	&  \cellcolor{gray!20}Tilted slightly to side (3) & \cellcolor{gray!20}Raised \\
\cellcolor{gray!5}Topic& \cellcolor{gray!5}Tilted slightly to side (3) & \cellcolor{gray!5}Raised, Widen eyes \\
\cellcolor{gray!20}Conditional	& \cellcolor{gray!20}Tilted slightly to side (3) &\cellcolor{gray!20}Raised  \\
 \hline
\end{tabular}}
\label{table:FE_list}
\end{table}

\begin{table}[!t]
\centering
\caption{ASL Grammar Errors Recognized by Our System.}
\parbox{0.4 \textwidth}{\small Lexical errors: if required facial expressions or head movements are not found within $S_L$ time from the sign. Timing errors (related to beginning and end): if facial expressions or head movements begin or end but with more than $S_{\text{BE}}$ time away from Clause Boundary.\\}
\resizebox{0.5 \textwidth}{!}{
\begin{tabular}{l  l  l }
  \hline
  \hline
Error Type &  Manual Sign &  Nonmanual Signal  \\

\hline
\cellcolor{gray!20}WHQ-Lexical & \cellcolor{gray!20}WHQ word & \cellcolor{gray!20}Not (WHQ or RHQ) \\

\cellcolor{gray!20}YNQ-Lexical & \cellcolor{gray!20}YNQ word & \cellcolor{gray!20}Not YNQ  \\

\cellcolor{gray!20}NEG-Lexical & \cellcolor{gray!20}Negative word & \cellcolor{gray!20}Not Negative  \\

\cellcolor{gray!20}COND-Lexical & \cellcolor{gray!20}Conditional word & \cellcolor{gray!20}Not Conditional \\ 
\hline
\hline
\cellcolor{gray!5}YNQ-Beginning & \cellcolor{gray!5}Clause Boundary &  \cellcolor{gray!5}YNQ begins \\ 

\cellcolor{gray!5}YNQ-End & \cellcolor{gray!5}Clause Boundary &  \cellcolor{gray!5}YNQ ends \\

\cellcolor{gray!5}COND-Beginning & \cellcolor{gray!5}Clause Boundary & \cellcolor{gray!5}Conditional begins  \\

\cellcolor{gray!5}TOPIC-Beginning & \cellcolor{gray!5}Clause Boundary &  \cellcolor{gray!5}Topic begins  \\
\hline
\end{tabular}}
\label{table:describ-errors}
\end{table}

\subsection{System Overview and Contributions}
\begin{figure*}[!t]
\includegraphics[width=\linewidth]{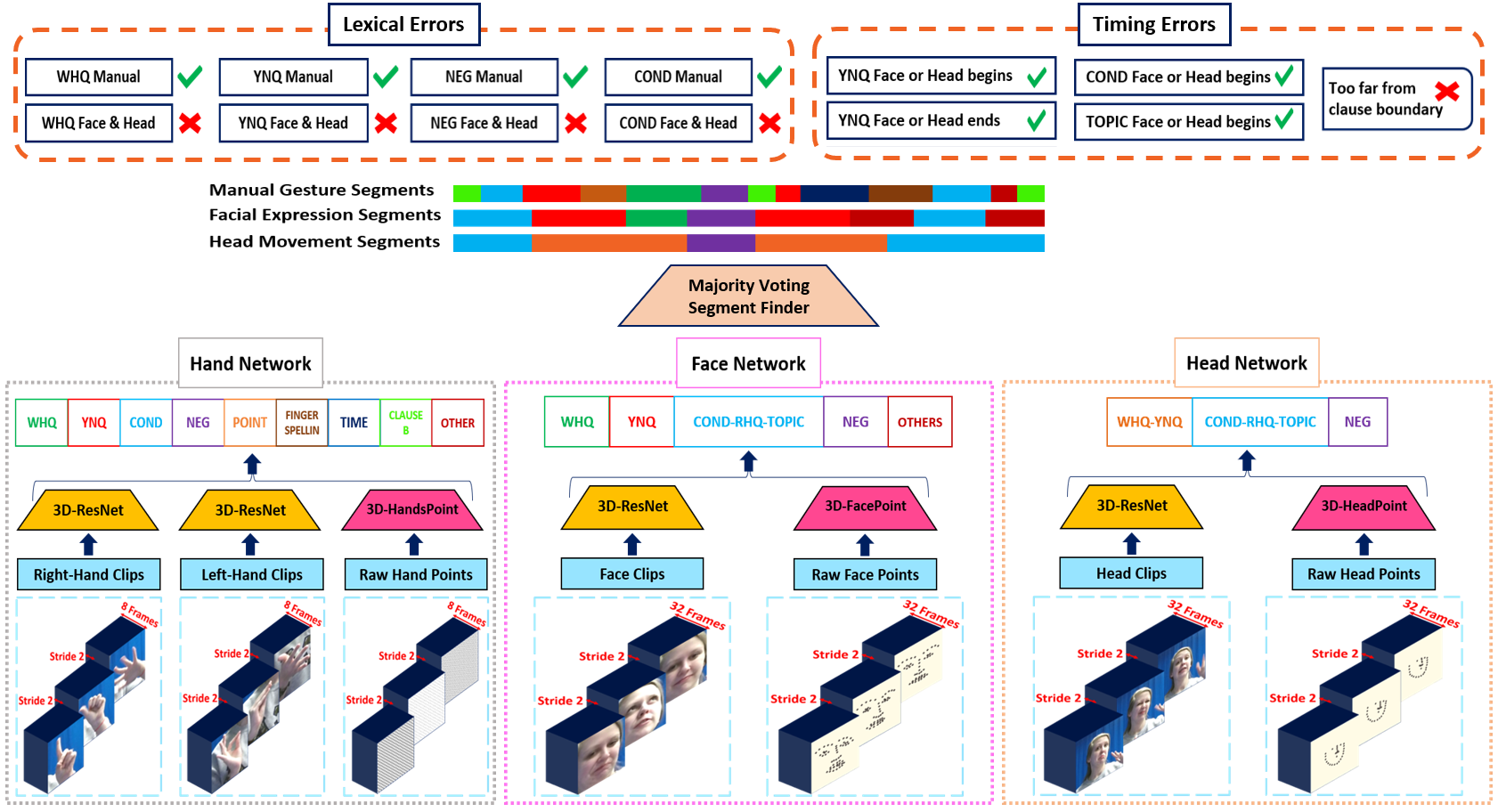}
\caption{The pipeline of the proposed multimodal 3DCNN framework for ASL nonmanual grammar recognition. The multiple modalities include hand gestures, facial expressions, and head movements. Raw coordinates and RGB clips of face and hands are fed to the networks to extract the pose and spatiotemporal information. Temporal boundaries of manual gestures and nonmanual signals are approximated using temporal sliding windows. The final predictions of frames are determined via Majority Voting algorithm and segments of all classes are extracted. Temporal correspondences of multiple modalities are compared to detect lexical and timing errors.}
\label{fig:framework}
\end{figure*}

As shown in Fig. \ref{fig:framework}, our framework includes 3D networks for multiple modalities: a \textit{Hand Gesture Network}, a \textit{Face Network}, and a \textit{Head Network} to capture spatiotemporal information from hands, facial expressions, and head movements respectively. Then a sliding window-based approach in temporal dimension is applied to detect temporal boundaries of grammatically important gestures. A majority voting algorithm is designed to finalize the predictions for each frame by picking the most frequent detected class of gesture. A segment detection algorithm is implemented to connect the predictions belonging to the same class. The predictions are then pruned based on the confidence scores returned by the networks. Finally, our \textit{Error Detection} algorithm detects grammatical mistakes by comparing the predictions from different modalities and analyzing their temporal correspondences.

The main contributions of the proposed framework are summarized as follows:

\begin{itemize}
\itemsep0em
    \item To the best of our knowledge, this is the first framework for automatic detection of ASL grammatical mistakes in continuous signing videos.   
    \item We propose a 3D multimodal framework to recognize grammatically important manual signs and nonmanual signals from continuous signing videos without temporal segmentation.  
    \item We have collected a continuous sign language dataset named as ASL-HW-RGBD, consisting of $1,026$ continuous videos of signing several sentences by fluent and student ASL signers. Our dataset covers different aspects of ASL grammar such as Conditional Sentences, Rhetorical Questions, WH Questions, YN Questions, Autobiography, Narrative, Pronouns, and Possessives. The dataset is annotated using ELAN annotation tool at the frame level. 
    \item Our system is able to recognize grammatical elements from manual gestures, facial expressions, and head movements respectively. Furthermore, by analyzing the temporal correspondence of the grammar elements from multiple modalities, our system can effectively and efficiently detect 4 lexical errors as well as 4 timing errors (related to beginning and end). 
    \item Our system generates instant feedback for ASL students about the detected grammatical mistakes in their continuous signing videos. 
\end{itemize}

\section{Related Work}

For continuous sign language recognition (CSLR), while most existing frameworks focus on the sentence-level translation of continuous signing, our system aims to recognize ASL grammatical mistakes without necessarily recognizing each word in a sentence. The existing CSLR frameworks consist of three main stages: 1) temporal segmentation, 2) feature extraction from segments, and 3) sequence modeling to reconstruct the sentence. Our approach is mostly related to the first two stages as we do not need to fully translate ASL sentences to find grammatical elements. 

\textbf{Temporal Segmentation}: Zhang \textit{et al.} proposed a threshold matrix-based method combined with hidden Markov model (HMM) for coarse segmentation, followed by dynamic time warping (DTW) for fine segmentation \cite{zhang2014threshold}.
Dawod \textit{et al.} used contrast adjustment and motion detection analysis for segmentation process \cite{dawod2016gesture}. Yang \textit{et al.} proposed dynamic time wrapping-based level building (LB-DTW) to segment sign sequences and then recognize signs \cite{yang2016continuous}. Connectionist temporal classification (CTC) \cite{graves2006connectionist} is an end-to-end sequence learning model which is suitable for unsegmented input data to learn the correspondence between the input and output sequences and is used in \cite{cui2017recurrent, pu2019iterative} for CSLR. Huang \textit{et al.} proposed a novel method based on hierarchical attention network with latent space to eliminate the need for temporal segmentation \cite{huang2018video}. Sliding window is a traditional approach generally used in dynamic event detection which is also employed for sign language recognition \cite{ong2014sign}. We employ a sliding window-based approach to detect temporal boundaries of manual gestures and nonmanual signals for grammar analysis. Sliding windows cover all segments even subtle facial movements but the temporal boundaries are roughly approximated.

\textbf{Video Feature Extraction}: 3D convolutional neural networks (CNNs) are commonly used to capture spatiotemporal representations in videos. 3DCNN was first proposed for video action recognition in \cite{3D}, and was improved in C3D \cite{C3D} by using a similar architecture to VGG \cite{VGG}. Recently, many 3DCNN models were proposed such as \cite{I3D},\cite{T3D},\cite{P3D} and demonstrated effectiveness in video recognition tasks. One of the main challenges of training deeper networks is gradient vanishing and gradient explosion which was tackled in ResNet\cite{he2016deep} by using \textit{skip connections} and sending the previous feature map to the next convolutional block. 3D-ResNet\cite{3DResNet} inherits the advantages of ResNet and also efficiently captures the temporal dimension. In this work, we employ 3D-ResNet \cite{3DResNet} as the backbone network for video feature extraction due to its capability to capture spatiotemporal features in action recognition tasks. 

\section{APPROACH}

We propose a multimodal framework to automatically recognize the grammatical errors from continuous signing videos without fully translating the entire sentences. Fig. \ref{fig:framework} summarizes the pipeline of the proposed framework. To analyze the grammar in continuous signing videos, we first recognize grammatical elements from manual gestures, facial expressions, and head movements by employing 3D-CNN networks to capture spatiotemporal features. We then analyze the correspondence of grammatical elements in different modalities and detect the correct and erroneous grammar-related signings. This section describes the data preprocessing, the design of hand gesture, face, and head networks in details, and our method for grammar error recognition. 

\subsection{Data Preprocessing} \label{section:Data Preprocessing}
We need to prepare the following components of the data for training: 1) Raw coordinates of face and hands key-points to extract pose information and dynamics of gestures. OpenPose \cite{cao2018openpose} is employed to extract raw coordinates of face and hands from RGB videos. If hands or face are out of frame boundary, OpenPose returns zero values for the key-points. In such cases, we estimate the key-points coordinates in the missed frames with interpolation and extrapolation of previous and next frames.  2) Regions of Hands and head to extract the fine-grained spatiotemporal information. The extracted coordinates by OpenPose are used to crop head and hands regions from RGB images.  
3) Registered face images to capture facial expressions without head movements.  We first calculate a \textit{mean face} by averaging the coordinates of facial landmarks across the entire dataset, then for each frame, we use the facial landmarks (extract by OpenPose \cite{cao2018openpose}) to register the face using affine transformation and with respect to the mean face. The raw face coordinates are also warped using the same process.

\subsection{Network Architecture} \label{section:Network-Architecture}

\text{\textbf{Temporal Boundary Recognition}}: To recognize temporal boundaries of gestures in continuous videos, most  existing methods used temporal segmentation of signs as a preprocessing step. However, incorrect temporal segmentation can consequently lead to wrong recognition of manual gestures and nonmanual signals thus negatively affect the entire pipeline by generating too many false positives or false negatives. Instead of using temporal segmentation, we add an extra  \textit{``Others''} class to the manual and facial gesture classes, which includes all instances that do not belong to any of the grammatically important categories. We then use a sliding-window approach to detect the intervals with high confidence scores of grammar elements. It is worth noting that this part is only used during testing where no temporal boundaries are provided. In the training phase, the annotations of temporal boundaries are used.

\textbf{Backbone network for video feature extraction:} We design three networks (Hand Gesture, Face, and Head) to extract video features. All of them use same backbone 3D-ResNet (with 34 layers) architecture with 5 convolutional blocks. The first one consists of one convolutional layer, followed by batch normalization, ReLU, and a max-pooling layer. The other 4 convolutional blocks are 3D residual blocks with skip connections. The number of kernels in the five convolutional blocks are \{$64, 64, 128, 256, 512$\}. The Global Average Pooling (GAP) is followed after the 5th convolutional block to produce a 512-dimensional feature vector. Then, one fully connected layer is applied to produce the final prediction. All the networks are optimized with Weighted Cross Entropy loss and Stochastic Gradient Descent (SGD) optimizer. 

\subsection{Hand Gesture Network} \label{section:Hand Gesture Network}

As described in Table \ref{table:manual_gesture_list}, the hand gesture network is trained to classify manual signs into 9 classes related to our system's error-detection rules: \textit{Conditional, Yes-No-Question (YNQ), WH-Question (WHQ), Negative, Time, Pointing, clause boundary, Finger-Spelling, and Others}. The class \textit{``Others''} includes all ASL signs that do not belong to the first 8 classes. The gesture network consists of three streams: RGB clip of right hand, RGB clip of left hand, and raw coordinates of hands. All streams are jointly trained in an end-to-end manner. The feature vectors from three streams are concatenated and fed to Softmax for classification. 

\textbf{3D-ResNets for Hands} extract spatiotemporal information of right hand and left hand clips by employing 3D-ResNet architecture with $34$ layers and temporal duration $8$, each 3D-ResNet producing a $512$-dimensional feature vector.

\textbf{3D-HandsPointNet} is designed to extract pose information and dynamics of hand gestures from raw coordinates of hands keypoints. It consists of two fully connected layers, each one followed by batch normalization and ReLU. The input is a $t \times 90$ matrix where $t$ is the temporal duration (set to $8$ frames in our system). Each column of the matrix is corresponding to one frame of the video and is a $90$-dimensional vector consisting of $x,y$ coordinates of $21$ key points of right hand, $21$ key points of left hand as well as the centers of right hand, left hand and face (to capture the relative positions). The coordinates are normalized using mean and standard deviation from the dataset. The dimension of the first and second fully connected layers are $90 \times 40$ and $40 \times 20$, respectively. The columns of the output matrix ($t \times 20)$ are concatenated resulting in a $160$-dimensional feature vector.

\subsection{Face Network} \label{section:Face Network}

As described in Table \ref{table:FE_list}, the face network is designed to classify facial expressions to 5 classes: \textit{Conditional-Topic-RHQ, Negative, Yes-No-Question (YNQ),  WH-Question (WHQ), and Others}. The \textit{Conditional-Topic-RHQ} class combines Conditional, Topic, and RHQ facial expressions as one class for classification due to their high similarity.  The class \textit{Others} includes all facial expressions that are not among the first 4 classes (not related to our system's error-detection rules). Face network consists of two streams: whole face region and coordinates of facial key points which are jointly trained and their feature vectors are concatenated. 

\textbf{3D-ResNet for Facial Expressions} receives RGB clips of registered face images as input resulting in a 512-dimensional feature vector. The purpose of face registration is to remove the head movements so that the network captures only facial expression changes. The temporal duration of this network is $32$ which is longer than Hand Gesture network because facial expressions typically last longer.

\textbf{3D-FacePointNet} takes clips of registered raw face coordinates as input, i.e., a $t \times 106$ matrix where $t$ is the temporal duration (set to $32$ in our system). Each column corresponds to one frame and is $106$-dimensional vector including $x,y$ coordinates of $53$ facial key points extracted by OpenPose \cite{cao2018openpose} (all the facial key points such as eyes, eyebrows, nose, mouth, except face contour). The network consists of two fully connected layers, each one followed by a batch normalization and a ReLU. The input matrix is spatially compressed to $t \times 20$ by the first fully connected layer. It is then transposed ($20 \times t$) and temporally compressed to $20 \times 10 $ by the second fully connected. The columns are then concatenated resulting in a $200$-dimensional feature vector. 

\subsection{Head Network} \label{section:head network}
The head network extracts spatiotemporal information of head movements. As listed in Table \ref{table:FE_list}, the head movements are categorized to three classes: \textit{Shaking head side-to-side (Negative), Head tilted forward (YNQ, WHQ), and Head tilted slightly to side (Topic, Conditional, RHQ)}. The head network also has two streams and is similar to face network with subtle differences.

The first stream is 3D-ResNet receiving clips of head images. The head regions are wider than face regions and are not registered. The second stream is 3D-HeadPointNet, receiving clips of raw coordinates of head as input which are $58$-dimensional vectors including 29 key points of face (face contour, nose, center of eyes and center of mouth) for each frame. 3D-HeadPointNet consists of two fully connected layers to reduce temporal and spatial dimensions and is similar to 3D-FacePointNet network. Likewise, the input from two streams are concatenated and fed to Softmax. 

\begin{figure}[!t]
\centering
\includegraphics[width=0.5\textwidth]{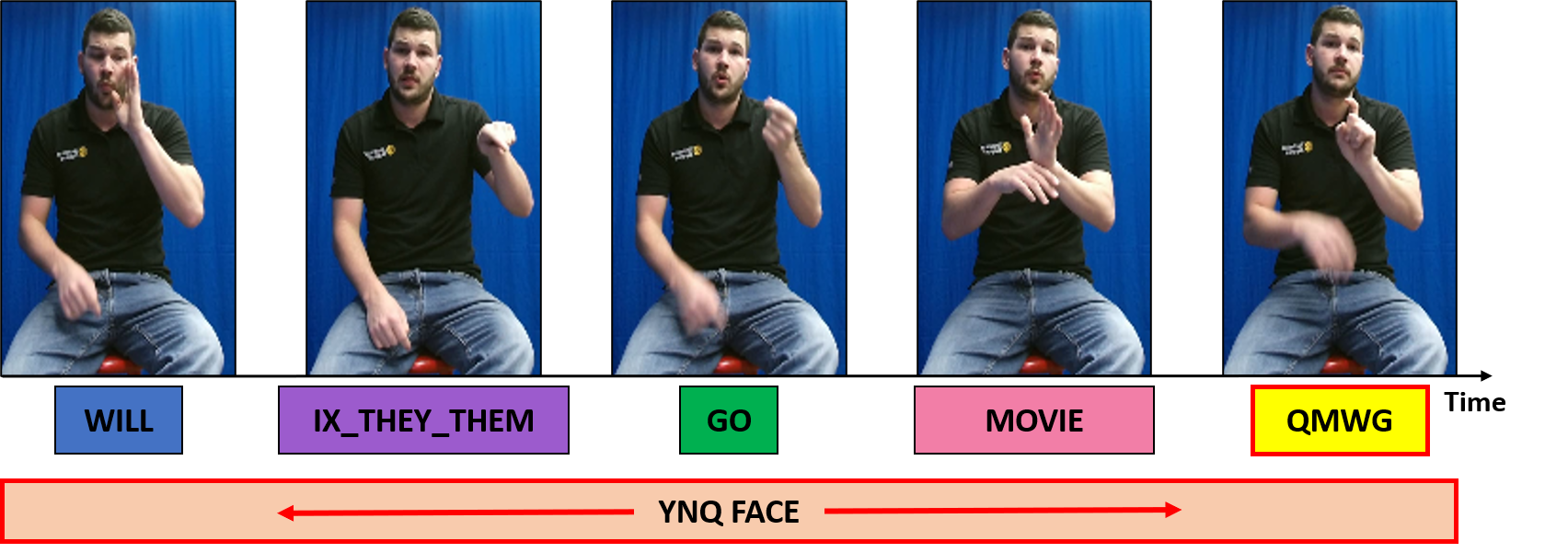}
\caption{An example of temporal alignment of signs in a Yes-No Question (YNQ) in ASL for the sentence ``Will they go to a movie?''. The YNQ facial expression happens during the entire question while the YNQ-related manual sign (QMWG) happens only at the end.
}
\label{fig:temporal-alignment}
\end{figure}

\begin{figure*}[!t]
\minipage{0.34\textwidth}
  \includegraphics[width=\linewidth]{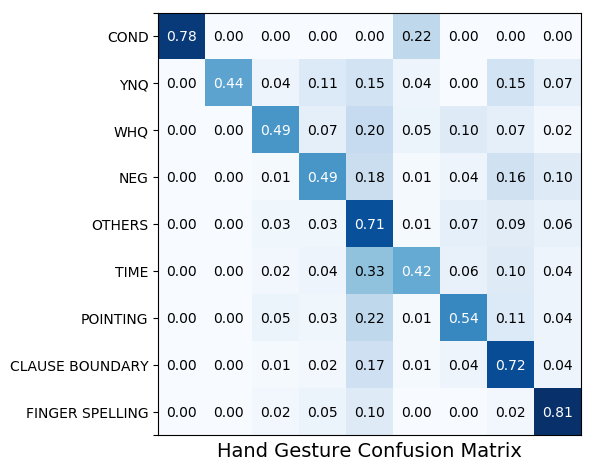}
\endminipage\hfill
\minipage{0.33\textwidth}
  \includegraphics[width=\linewidth]{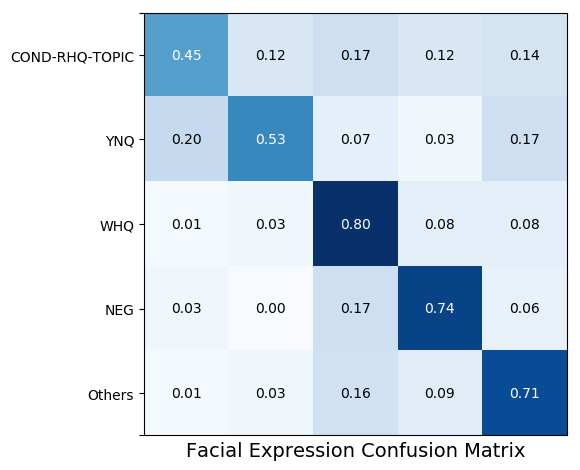}
\endminipage\hfill
\minipage{0.33\textwidth}%
  \includegraphics[width=\linewidth]{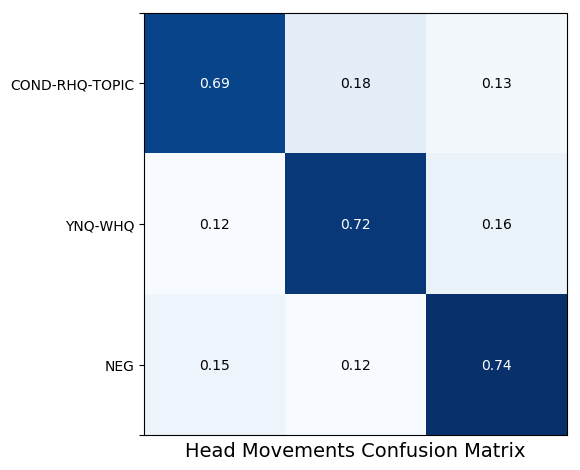}
\endminipage
\caption{Normalized confusion matrices of Hand Gesture Network with $66.84\%$ accuracy (left), Face Network with $66.32\%$ accuracy (middle), and Head Network with $71.14\%$ accuracy (right). The rows are ground truth labels and the columns are predictions.
}
\label{fig:confusion-matrix-results}
\end{figure*}

\subsection{Grammatical Error Recognition} \label{section:Grammatical Error Recognition}
Our method for recognizing ASL grammar errors contains two main steps: ASL segment detection and grammatical error recognition.
Note that the grammatical error recognition is only conducted in testing phase.

\textbf{ASL Segment Detection:} In order to detect the temporal boundaries of gestures, a sliding window is applied throughout the input video, and the frames of each window are fed to the networks. In our experiments, the size of sliding window is 8 for hand gesture network and 32 for face and head networks with a stride at 2. Denoting the size of sliding window by $S$, there are $\ceil*{\frac{n-S+1}{2}}$ sliding windows for a video with $n$ frames. To speed-up the process, at each step, a batch of 64 sliding windows are fed to the network in parallel for testing. 
 
Each video frame belongs to several overlapping sliding windows resulting in a list of predictions for each frame. A Majority Voting method is implemented to determine the final prediction for each frame, i.e. by selecting the most frequent predicted labels in the list and referring to prediction of previous frames if there is a tie. The pseudocode is illustrated in \textit{Algorithm \ref{majority-alg}}. 

\begin{algorithm}
\caption{\textit{Majority Voting Algorithm}}\label{euclid}
\begin{algorithmic}[1]
\Procedure{}{}
\State $i \gets \text{current frame index}$
\State $\textit{P} \gets \text{final predictions for previous frames}$
\State $\textit{l} \gets \text{list of predictions for frame } i$
\State $T \gets \text{empty hash table}$
\BState \emph{loop over $l$}:
\If {$\text{class } C \in \text{keys} $} $T[C] +=1 $
\Else \ $T[C]=1$
\EndIf
\BState \emph{candidates}:
\State $K= [ \ C \in T.\text{keys \ if \ } T[C]= \max(T.\text{values}) \ ]$
\If {$|K|=1$} \Return $K[0]$
\EndIf 
\State $j \gets 1$
\While{$j \leq 3 $}
\If {$P[i-j] \in K$} \Return $P[i-j]$
\EndIf
\State $j \gets j+1$
\EndWhile
\State \Return \text{a random element of }$K$
\EndProcedure
\end{algorithmic}
\label{majority-alg}
\end{algorithm}

After obtaining the final prediction for each frame via Majority Voting algorithm, consecutive frames with the same label (i.e. segments) are detected from manual gestures, facial expressions, and head movements independently. For the frames in each segment with the same prediction, if confidence score of at least one frame is greater than the specified threshold, the prediction is kept without change. Otherwise, the predicted label is changed to class \textit{Others}. In our experiments, the confidence score threshold is set to $0.8$ for all modalities.

\textbf{ASL Grammatical Error Recognition:} ASL grammar errors are recognized based on a set of rules related to the synchronization between the manual gestures, facial expressions, and head movements, as listed in Table \ref{table:describ-errors}. The facial expressions (and head movements) in ASL are usually longer than the manual gestures, making the temporal alignments more complicated. Fig. \ref{fig:temporal-alignment} shows an example of temporal alignment for a Yes-No Question (YNQ) while the signer is performing a sentence consisting of five manual signs: ``WILL,'' ``IX-THEY-THEM,'' ``GO,'' ``MOVIE,'' and ``QMWG'' (question mark wiggling sign). Rather than performing the YNQ nonmanual signal only during QMWG, the signer performs it during the entire sentence (which is also grammatically acceptable). Our system looks for potential lexical and timing errors (see definitions in Table \ref{table:describ-errors}), and sets temporal proximity thresholds of $200$ msec for lexical errors and $1$ second for timing errors.

\section{Continuous Sign Language Dataset} \label{dataset-section}

We have collected a sign language dataset, ASL-HW-RGBD, consisting of $1,026$ continuous videos (some consisting of multiple ASL sentences), as produced by 46 ASL signers, including both fluent and student signers. Each video was produced as a response to a homework-style assignment. Some assignments simply ask students to translate a set of English sentences into ASL, and others ask students to invent short multi-sentence passages. Each assignment is designed to elicit sentences that consist of several ASL grammar elements: For instance, the first homework is mainly about signing two basic question types in ASL, which are WH-Questions (what, who, where, how, etc) and Yes-No-Questions. The second homework asks students to compose longer multi-sentence videos with autobiographical details. The third homework encourages students to use various pointing signs and fingerspelling, while the forth homework encourages students to create a longer video with multiple items under discussion and questions asked to the camera. The fifth homework is centered around pronouns, negation, and questions. Finally, the sixth homework is about conditional sentences and rhetorical questions.

The dataset has been annotated using the ELAN annotation tool, with several parallel timeline tiers of annotations for each video (in which spans of time in the video are labeled with key linguistic information). We briefly list some of the most important tiers for our research: 1) Clause: beginning and end of the clause, 2) Fingerspelling: spelling out words by using hand shapes that correspond to the letters of the word, 3) Lexical Pointing: referring to person or concept under discussion, 4) Wanted Words: words that may be used more frequently than others including several classes such as WH-Question, Yes-No Question, Negative, Conditional, Time, and Pointing, 5) Facial Expressions including several classes such as Conditional, Negative, Yes-No Question, WH-Question, Rhetorical Question, and Topic.

\section{EXPERIMENTS}

\subsection{Implementation Details}

Our proposed models are implemented in PyTorch on four Titan X GPUs. To avoid over-fitting, our models are pretrained on Kinetics \cite{Kinetics} which is a large human action dataset. The original resolution of RGB videos in our ASL-HW-RGBD dataset is $1,920 \times 1,080$ pixels. To prepare the input for hand gesture and head networks, we use the coordinates extracted by OpenPose \cite{cao2018openpose} to crop $240 \times 240$ regions around the center of face and $160 \times 160$ regions around the center of each hand. The $240 \times 240$ face images are used for face registration, and are further cropped to $100 \times 128$ for face network. All these bounding boxes are adjusted based on the mean and variance of key-points coordinates in the dataset to guarantee the face regions are inside the 100x128 bounding box.  We finally resize all input images to $134 \times 134$ for training and testing. Data augmentation techniques such as random cropping and random rotation are used during training. In every iteration of training, $112 \times 112$ image patches are randomly cropped from $134 \times 134$ input images for data augmentation. Random rotation (with a degree randomly selected in a range of [$-10$, $10$]) is applied on the cropped patches to further augment the data. During the testing, only the center patches of size $112 \times 112$ are used for predictions. The models are fine-tuned for $\mathit{200}$ epochs with an initial learning rate of $\lambda = 3\times10^{-3}$, reduced by a factor of $\mathit{10}$ every $50$ epochs. The batch size is $128$.

\subsection{Grammar Element Detection Results}

The grammar error recognition relies on the detection accuracy of the manual signs and nonmanual signals. A wrong prediction of manual gestures consequently leads to search for a non-essential element in other modalities, resulting in a false positive grammar error. Moreover, bias towards predicting a certain class of face or head can lead to missing grammatical mistakes when that class is required but has not been performed, resulting in a false negative detection.

Normalized confusion matrices of Hand Gesture, Face, and Head Networks are shown in Fig. \ref{fig:confusion-matrix-results}. The value in cell $(i,j)$ of each matrix is the percentage of predicting ground truth label $i$ as class $j$. We observe that misclassifications are mostly caused by class \textit{Others}. The reason is that class \textit{Others} is heterogeneous including all instances that do not belong to rest of categories, but may contain similar manual gestures and nonmamual signals. Therefore, it is easy to be confused with the rest of classes.
 
\subsection{Grammatical Error Recognition Results}

To evaluate our framework, we test a subset of videos in our ASL-HW-RGBD dataset with labeled grammatical errors listed as Table \ref{table:describ-errors}. The grammatical errors in these clips are detected and annotated by ASL linguists beforehand. Note that in the training phase, the input data does not include grammatical errors at all. For testing, we first recognize the grammatical elements from multiple modalities and then analyze grammar errors independently. The lexical errors happen if a grammatically important hand gesture is recognized with more than $0.8$ confidence score but the corresponding facial expression and head movement are not found within $200$ msec from the hand gesture. The timing errors happen if there is a grammatically important facial expression or head movement (with larger than $0.8$ confidence score) but it is performed more than $1$ second away from the closest clause boundary. The results of grammatical error recognition are shown in Table \ref{tab:GT-errors}. The second column includes the ground truth number of instances and the third column is the number of instances recognized by our system. The last column (True Positive Rate) is the ratio of the recognized errors to ground truth errors. For \textit{Error-COND-Beginning}, \textit{Error-YNQ-Beginning} and \textit{Error-YNQ-End}, there are not enough ground truth instances so they are not included in the evaluation. The average true positive rate is $60\%$. The true positive rate of \textit{Error-COND-Lexical} is the highest among lexical errors because ``Conditional'' hand gestures can be recognized with relatively high accuracy ($78\%$) compared to other classes. On the other hand, the recognition accuracy of ``YNQ'', ``WHQ'', and ``Negative'' hand gestures are less than $50\%$ (see Fig. \ref{fig:confusion-matrix-results}) and can be missed in some cases resulting in false negative error detection. Furthermore, the true positive rate of \textit{Error-TOPIC-Beginning} is the highest because recognition of the ``Topic'' class either from face or head leads to exploring the possibility of this error and with relatively high recognition accuracy of ``Clause Boundary'' from hand gestures, this error can be correctly recognized in most cases. 

Our system is able to generate feedback for students on average in less than 2 minutes for 1 minute ASL videos including all the steps from pre-processing the video, testing the networks, and error detection.

\begin{table}[!htb]
\centering
\caption{ASL nonmanual grammatical error recognition results. }
\resizebox{0.5\textwidth}{!}{
\begin{tabular}{|l |c| c|c| }
  \hline
  Error Type  & Ground Truth & Recognized & TP Rate \\
  \hline
  \hline
\cellcolor{gray!10}Error-YNQ-Lexical	& \cellcolor{gray!50} $12$  &\cellcolor{gray!10} $7$ & \cellcolor{gray!50} $58.3\%$ \\
\cellcolor{gray!10}Error-NEG-Lexical	& \cellcolor{gray!50} $36$ &\cellcolor{gray!10} $14$ & \cellcolor{gray!50} $38.8\%$ \\
\cellcolor{gray!10}Error-WHQ-Lexical	& \cellcolor{gray!50} $40$ &\cellcolor{gray!10} $23$ & \cellcolor{gray!50} $57.5\%$ \\
\cellcolor{gray!10}Error-COND-Lexical	&  \cellcolor{gray!50}$24$ &\cellcolor{gray!10} $18$ & \cellcolor{gray!50} $75.0\%$ \\
\cellcolor{gray!10}Error-TOPIC-Beginning& \cellcolor{gray!50} $23$ &\cellcolor{gray!10} $19$ & \cellcolor{gray!50} $82.6\%$ \\
\hline
\cellcolor{gray!10} Total	&  \cellcolor{gray!50}    $135$ &\cellcolor{gray!10} $81$ & \cellcolor{gray!50} $60.0\%$ \\
  \hline
\end{tabular}}
\label{tab:GT-errors}
\end{table}

\section{CONCLUSION}

In this paper, we have proposed a 3DCNN-based multimodal framework to automatically recognize ASL grammatical elements from continuous signing videos. Then a sliding window-based approach is applied to detect 8 types of ASL grammatical mistakes by checking the temporal correspondence between manual signs and nonmanual signals in signing videos. Our system generates instant feedback for ASL students about the grammatical aspects of their performance without fully translating the sentences. We have collected and annotated a new dataset, \textit{ASL-HW-RGBD}, consisting of $1,026$ continuous sign language videos. Our system is able to recognize $60\%$ of grammatical mistakes. Our future work will aim to develop more advanced methods to handle the complex relations between manual gestures and nonmanual signals in continuous ASL videos to improve the recognition accuracy of ASL grammatical elements as well as ASL grammar errors.

\section*{ACKNOWLEDGMENT}
This material is based upon work supported by the National Science Foundation under award numbers 1400802, 1400810, and 1462280.



\end{document}